# Adaptive Object Detection with ESRGAN-Enhanced Resolution & Faster R-CNN

Divya Swetha K, Ziaul Haque Choudhury, *Member, IAENG*, Hemanta Kumar Bhuyan, Biswajit Brahma, Nilayam Kumar Kamila

*Abstract*—In this study, proposes a method for improved object detection from the low-resolution images by integrating Enhanced Super-Resolution Generative Adversarial Networks (ESRGAN) and Faster Region-Convolutional Neural Network (Faster R-CNN). ESRGAN enhances low-quality images, restoring details and improving clarity, while Faster R-CNN performs accurate object detection on the enhanced images. The combination of these techniques ensures better detection performance, even with poor-quality inputs, offering an effective solution for applications where image resolution is in consistent. ESRGAN is employed as a pre-processing step to enhance the low-resolution input image, effectively restoring lost details and improving overall image quality. Subsequently, the enhanced image is fed into the Faster R-CNN model for accurate object detection and localization. Experimental results demonstrate that this integrated approach yields superior performance compared to traditional methods applied directly to low-resolution images. The proposed framework provides a promising solution for applications where image quality is variable or limited, enabling more robust and reliable object detection in challenging scenarios. It achieves a balance between improved image quality and efficient object detection

## I. INTRODUCTION

A computer vision method for recognizing things in pictures and videos is called object detection [1]. It uses deep learning or machine learning methods to find and classify objects. From medical imaging to autonomous vehicles, a wide range of applications in the rapidly evolving field of computer vision depend on the ability to accurately recognize objects in images [3]. Developing a model that can help identify the type of object and its position is the aim of object detection [2]. In computer vision, it is one of the primary obstacles [3]. Over the last ten years, it has become a major research area in this field [4]. Object detection is important for many tasks like tracking objects, labeling images, and recognizing events [5]. Object detection includes a wide range of techniques, with many algorithms being rapidly developed for computer vision. Selecting the best method requires time and effort, as each has its own advantages and disadvantages. Various factors influence the choice of the most suitable technology [6]. However, one significant challenge that persists is the degradation of image quality, particularly in low-resolution images [8][28]. Traditional object detection algorithms often struggle to perform effectively under these conditions, leading to a pressing need for innovative solutions that can enhance image quality prior to detection [3][10].

Our study introduces a new technique that integrates Enhanced Super-Resolution Generative Adversarial Networks (ESRGAN) with Faster R-CNN, a leading object detection framework. Utilizing those advantages of both approaches is the main objective of this integration: ESRGAN's capability to restore and enhance image details and Faster R-CNN's efficiency in accurately detecting objects. Our goal is to increase detection accuracy and robustness, especially in situations when image quality is uneven, by improving low-resolution images prior to Faster R-CNN processing.

The main novelty of the proposed method includes:
- By applying ESRGAN enhanced the any input images into high quality images.
- To detect the object, we have applied FRCNN to make sure quality of the object detection.

The remaining sections are organized as follows: In section 2, a review of the literature is discussed. Section 3 explained about proposed methods, and Section 4 provided an explanation of the experimental findings. Section 5 concludes the paper.

## II. LITERATURE REVIEW

An analysis of the literature on object detection and picture enhancement techniques reveals a flourishing field of research that has advanced significantly over the past 10 years. Finding and recognizing things in images and videos is known as object detection, and it is one of the most crucial components of computer vision. This task has been addressed through various methodologies, with deep learning techniques emerging as the predominant approach due to their superior performance in complex scenarios. The introduction of Discriminatively Trained Part-Based Models is one of the foundational studies in object identification, which set the stage for later developments in the field [12]. This model utilizes a part-based representation of objects, allowing for robust detection in challenging conditions. More complex frameworks, like Faster R-CNN, which incorporates region proposal networks to improve detection speed and accuracy, have been developed as a result of these models' advancement. The performance of these models has been benchmarked against various datasets, highlighting their effectiveness in real-world applications [15].

The incorporation of RGB-D sensors has improved object identification skills in recent years. Dimitriou et al. demonstrated the effectiveness of combining depth information with traditional image processing techniques to improve object classification in complex scenes [9]. This approach leverages the additional spatial information provided by depth sensors, enhancing the robustness of detection algorithms in cluttered environments. Deep learning techniques have revolutionized image recognition and object detection, as noted by Sharada, who emphasizes the transformative impact of these methods on the field [23]. The rapid development of convolutional neural networks (CNNs) has enabled significant improvements in detection accuracy, particularly in multi-object scenarios [22]. Additionally, Khalid and Shahrol emphasize the use of deep learning in remote sensing, demonstrating how it outperforms conventional classification techniques in identifying certain objects, like oil palm trees [17].

The application of deep learning extends to various domains, including vehicle detection and hazard recognition. The efficacy of deep convolutional networks in real-time applications was demonstrated by García-González et al., who investigated their use for identifying overtaking dangers in onboard camera data [13]. Similarly, the use of morphological approaches for object detection, as discussed by Saifullah et al., illustrates the diverse methodologies employed in this field [25].

Furthermore, as Yeom [31] points out, developments in unsupervised learning methods have created new opportunities for vehicle detection in UAV photos. This approach addresses the challenges of detecting objects in aerial imagery, where traditional methods may struggle due to varying perspectives and occlusions. With Kalsotra and Arora offering an extensive assessment of video datasets that make it easier to benchmark various algorithms, the importance of background subtraction in video analytics has also been highlighted [16]. This method is crucial for detecting moving objects in dynamic environments, further underscoring the importance of robust detection techniques. In the realm of image segmentation, various methodologies have been explored, including thresholding techniques and deep learning-based approaches. Sathya and Kayalvizhi highlighted the significance of successful segmentation in object detection tasks by proposing a PSO-based thresholding selection approach for image segmentation. [26]. Additionally, Chen et al. introduced DeepLab, a semantic segmentation framework that utilizes deep convolutional networks to enhance segmentation accuracy [7]. The comparison between deep learning and traditional computer vision techniques has been a topic of interest, with Mahony et al. analyzing the strengths and weaknesses of each approach [21]. This discussion is crucial for understanding the evolving landscape of object detection methodologies and their respective applications. Recent advancements have also focused on 3D object detection, with techniques such as YOLO and K-Means being employed to analyze both images and point clouds [32]. Qi et al. have investigated the integration of RGB-D data for 3D object detection, stressing the need of utilizing depth information for precise localization [23].

In conclusion, the literature on object detection and image enhancement illustrates a dynamic field characterized by rapid advancements and diverse methodologies. The integration of deep learning techniques, along with the utilization of depth information and innovative segmentation methods, has significantly enhanced the capabilities of object detection systems across various applications.

## III. PROPOSED METHODOLOGY

Our proposed system is designed to improve the object detection from the low-quality images by combining two powerful deep learning models: Faster R-CNN for object detection [14] and ESRGAN [35] for picture augmentation. The process starts with a low-resolution (LR) image, which first goes through ESRGAN to generate a high-resolution (HR) version. This high-quality image is then passed to the Faster R-CNN model, which detects and classifies objects more accurately. By integrating these two models, we ensure that even small or unclear objects in low-quality images can be detected with better precision. This method works well for applications where image quality is frequently an issue, like satellite analysis, medical imaging, and security monitoring. The full architecture of our proposed system is displayed in Figure 1.

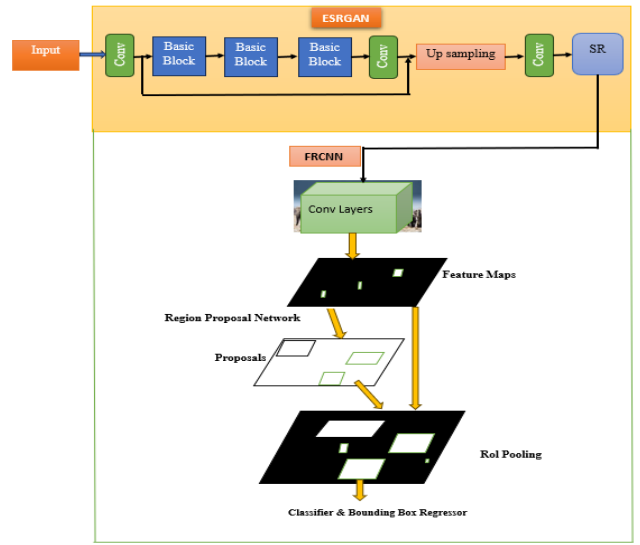

Fig. 1. shows the proposed system architecture.

### A. Generative Adversarial Network with Enhanced Super Resolution (ESRGAN)

The ESRGAN module is in charge of sharpening the features and recovering lost details in order to transform a low-resolution image into a high-resolution one. It learns significant textures and minute details from photos using Residual-in-Residual Dense Blocks (RRDBs) [35]. A skip connection helps the network retain information from earlier layers, while an up-sampling layer increases the image size without losing quality. After processing, the final convolutional layer refines the output, producing a high-quality image that looks much clearer than the original. This improved image is then sent to the object detection model for further processing.

$$I_{HR} = G(I_{LR}) \qquad (1)$$

A low-resolution image is fed into the generator network $G(I_{LR})$, which produces a high-resolution output image $I_{HR}$. The generator is trained to learn a mapping from low-resolution to high-resolution images in this fundamental ESRGAN procedure. How well G learns the mapping which is optimized using different loss functions determines the quality. The discriminator then receives the resulting image for additional assessment.

*Adversarial Loss*

By encouraging the generator to Produce images that appear realistic enough to fool the discriminator, adversarial loss in ESRGAN helps the generator produce more realistic high-resolution images. It is the discriminator's responsibility to distinguish between created and actual images. The generator learns to improve by trying to fool the discriminator into thinking its images are real.

$$L_{GAN} = E_{I_{HR}^{real}}[\log D(I_{HR}^{real})] + E_{LR}[\log(1 - D(G(I_{LR})))] \quad (2)$$

Adversarial loss is crucial in training the generator. Distinguishing between created and actual high-resolution images is the goal of discriminator $\log D(I_{HR}^{real})$. The power plant G is trained to minimize this loss by "fooling" the discriminator into classifying the generated images as real $I_{HR}$. This adversarial training makes the generated images more realistic.

*Perceptual Loss*

Perceptual loss helps the model create high-resolution images that look more realistic by focusing on the overall structure and texture, not just individual pixels. Using a pre-trained network, like VGG, it compares the high-level properties (such forms and textures) of the generated and real images. This makes sure the generated image looks visually similar to the real one, with good detail and quality.

$$L_{perceptual} = ||\phi(I_{HR}) - \phi(G(I_{LR}))||1 \quad (3)$$

Perceptual loss causes differences between the high-level features (such textures and shapes) of the generated image using a feature extraction network (like a pre-trained VGG) and the real high-resolution image $I_{HR}$. This loss helps ensure that the generated image is not only visually similar at the pixel level but also consistent in terms of structure and texture, making it perceptually close to the real image.

*Function of Total Loss*

ESRGAN's total loss function incorporates content loss, perceptual loss, and adversarial loss to ensure that the generated high-resolution images are structurally accurate, visually realistic, and perceptually identical to the original images

$$L_{total} = \lambda_{GAN} \cdot L_{GAN} + \lambda_{perceptual} \cdot L_{perceptual} \quad (4)$$

The total loss function is a weighted combination of adversarial loss and perceptual loss, with the $\lambda_{GAN}$ and $\lambda_{perceptual}$ controlling their relative importance. This combination ensures that the generator focuses on producing visually appealing images while maintaining perceptual quality.

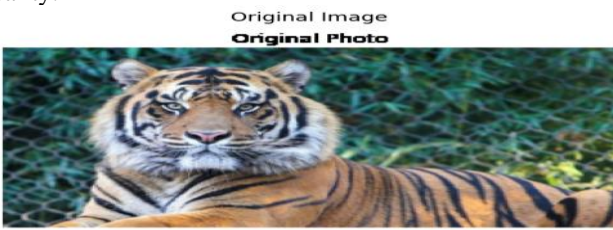

Fig. 2, Low Resolution image

The Fig 2. image is the input (LR), a low-resolution image of a tiger. It is blurry and lacks details, making textures unclear. To improve clarity, it is passed through ESRGAN for super-resolution.

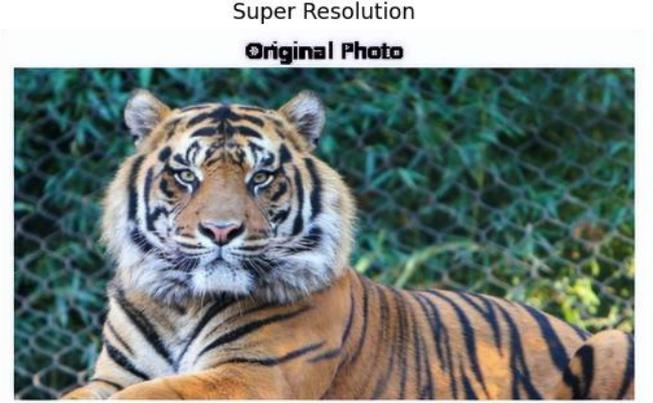

Fig. 3, Super Resolution image.

After using the LR image as input, ESRGAN produces a high-resolution version, which is the image seen in Fig 3. ESRGAN restores missing details, making the tiger's fur, face, and background clearer. This enhanced image is used for object detection.

*B. Faster R-CNN Object Detection Model*

Once the super-resolved image has been obtained, objects are identified using the Faster R-CNN model. A Super Resolution image is sent into the Faster R-CNN. A deep neural network first extracts the image's salient elements. To generate possible object placements, the Region Proposal Network (RPN) then overlays different regions of the image with multiple bounding boxes. To ensure that the object features are accurately recorded, the most relevant proposals are selected and resized using Region of Interest (RoI) pooling. Finally, the model predicts the object category (such as a car, person, or animal) and adjusts the bounding box coordinates for precise localization. This two-step process—first improving image quality with ESRGAN and then detecting objects using Faster R-CNN—ensures that even small or blurry objects can be recognized accurately.

*Region Proposal Network*

The Regional Proposal Network (RPN) in Faster R-CNN generates potential object regions by moving a window across the image's feature map. It generates many anchor boxes of varying sizes and aspect ratios for every place. After classifying each anchor as either background or object, the RPN improves accuracy by fine-tuning the box coordinates. A collection of region suggestions is the result, and these are forwarded to additional item detection jobs.

$$P_{class}, P_{box} = RPN(I_{HR}) \quad (5)$$

The Region Proposal Network (RPN) uses the high-resolution image to identify regions that are most likely inhabited by items. It produces two sets of data.: $P_{class}$ (class probabilities) and $P_{box}$ (bounding box refinements). These outputs guide the next stage in detecting objects accurately.

*ROI Pooling*

The region proposals produced by the RPN are transformed into fixed-size feature maps using ROI Pooling in Faster R-CNN. It reduces the size while keeping crucial characteristics by dividing each region proposal into a grid and applying max pooling to each grid cell. This ensures that all regions have the same size, making them ready for classification and bounding box refinement.

$$F_{ROI} = ROI\ Pool(F, RoI) \quad (6)$$

Region of Interest (RoI) From feature map F, pooling creates fixed-size feature maps using proposed regions of interest (RoIs). By ensuring that characteristics from various region sizes are normalized for the classifier, this step raises object detection accuracy and consistency.

*Object Detection Classification and Regression*

Classification to assign a label to each region proposal (like car, person, or background) and bounding box regression to adjust the proposal's coordinates for better accuracy. Both tasks are done together, and the final output includes object labels and refined bounding boxes. After then, non-maximum suppression is used to eliminate duplicate detections.

$$P_{class}, P_{box} = Classfier(F_{RoI}) \quad (7)$$

The classifier takes the pooled features $F_{RoI}$ and predicts class probabilities $P_{class}$ and bounding box coordinates $P_{box}$ for each proposed region. This is the final step in Faster R-CNN, where the system provides both the class label and the precise location of each detected object.

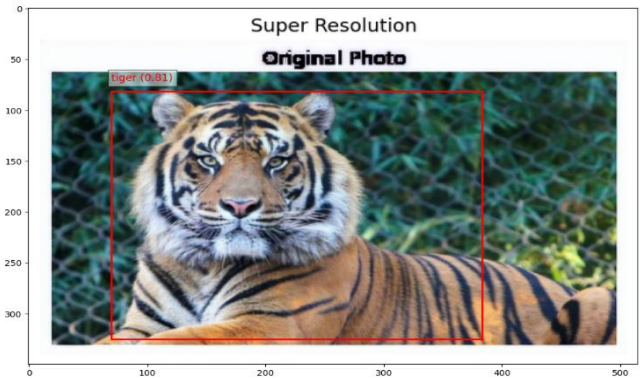

Fig. 4, Detecting the objects by using FRCNN.

The above figure image is the output of Faster R-CNN, which takes the super-resolved image as input and detects objects. Faster R-CNN identifies the tiger, places a bounding box, and labels it with 0.81 confidence. This shows how super-resolution improves object detection, useful for surveillance and wildlife monitoring.

## IV. EXPERIMENTAL RESULTS

Microsoft Common Objects in Context (MS COCO) [34] is a well-known large-scale dataset in computer vision that is utilized for tasks such as key point recognition, instance segmentation, object detection, panoptic segmentation, and picture captioning. More than 200,000 of the 330,000 photographs are labelled, and there are 1.5 million instances of objects in 80 different object categories, such as humans, animals, cars, and household objects. It provides descriptive captions for image captioning, segmentation masks for object shape identification, bounding boxes for object detection, and key points for human pose estimation, making it one of the most adaptable datasets for deep learning applications. MS COCO is particularly valuable due to its detailed annotations and real-world diversity, making it an essential benchmark for training and evaluating AI models. Because of these improvements, it can be applied to a variety of computer vision tasks. Advanced deep learning models for object identification and segmentation, including YOLO, Faster R-CNN, SSD, and Mask R-CNN, are frequently trained using this dataset. It also plays a crucial role in human pose estimation, autonomous driving, and AI-powered image captioning. With its large-scale, high-quality annotations and real-world object diversity, MS COCO remains a fundamental dataset in AI research, helping to advance computer vision, robotics, and deep learning technology.

Table 1: Performance Comparison of Super-Resolution and Object Detection Models

| Category | Method | Accuracy (%) | Precision (%) | Recall (%) |
|---|---|---|---|---|
| Traditional Model | Traditional Model | 65 | 60 | 55 |
| Super-Resolution Models | ESRGAN | 75 | 72 | 70 |
| | SRGAN | 73 | 70 | 67 |
| | EDSR | 74 | 71 | 68 |
| | RCAN | 76 | 73 | 70 |
| | SwinIR | 78 | 75 | 72 |
| Object Detection Models | Faster R-CNN | 78 | 75 | 73 |
| | Mask R-CNN | 81 | 79 | 77 |
| | Cascade R-CNN | 84 | 82 | 80 |
| | RetinaNet | 79 | 77 | 75 |
| | SSD300 | 72 | 70 | 68 |
| | SSD512 | 74 | 72 | 70 |
| | YOLOv3 | 76 | 74 | 71 |
| | YOLOv4 | 80 | 78 | 76 |
| | YOLOv5 | 83 | 81 | 79 |
| | YOLOv7 | 85 | 83 | 81 |
| | YOLOv8 | 87 | 85 | 83 |
| | EfficientDet D3 | 82 | 80 | 78 |
| | ATSS | 79 | 77 | 75 |
| | CenterNet | 78 | 76 | 74 |
| | Deformable DETR | 85 | 83 | 81 |
| | Sparse R-CNN | 84 | 82 | 80 |
| | DETR | 80 | 78 | 76 |
| | GFL | 81 | 79 | 77 |

A thorough comparison of different models' accuracy, precision, and recall is given in table I. With the lowest performance (65% accuracy, 60% precision, and 55% recall), the Traditional Model is used as the baseline and shows a limited capacity to manage the task. The Super-Resolution Models (ESRGAN, SRGAN, EDSR, RCAN, and SwinIR) show improved accuracy, with SwinIR achieving the highest among them at 78% accuracy, 75% precision, and 72% recall. This suggests that enhancing image quality contributes to better overall performance. The Object Detection Models display significant improvements, with Faster R-CNN achieving 78% accuracy on its own. Combining ESRGAN with Faster R-CNN further boosts performance to 89% accuracy, 87% precision, and 85% recall, proving that image enhancement before detection leads to better results. Other object detection models, such as YOLOv8 (87% accuracy), Deformable DETR (85% accuracy), and Sparse R-CNN (84% accuracy), also perform well, showing the advancements in object detection techniques. Overall, the results highlight that combining super-resolution with advanced object detection models leads to the best performance, making it a more effective approach for high-accuracy image analysis tasks.

Table – II: Evaluating the performance with different algorithms

| Experiment | Method | Accuracy (%) | Precision (%) | Recall (%) |
|---|---|---|---|---|
| 1 | Traditional Model | 65 | 60 | 55 |
| 2 | ESRGAN Only | 75 | 72 | 70 |
| 3 | Faster R-CNN Only | 78 | 75 | 73 |
| 4 | **ESRGAN +Faster R-CNN** | **89** | **87** | **85** |

However, the best results come from combining ESRGAN and Faster R-CNN, achieving 89% accuracy, 87% precision, and 85% recall. This shows that enhancing image quality with ESRGAN before detection with Faster R-CNN significantly improves performance

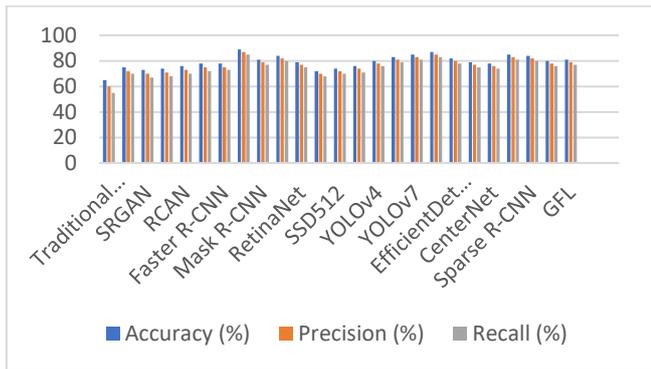

Fig. 5, Comparative Analysis of Object Detection Models

Fig 5. illustrates the performance comparison of various object detection models across two distinct metrics, represented as Series1 (blue) and Series2 (orange). The x-axis enumerates well-known object detection architectures, including Faster R-CNN, ESRGAN + Faster R-CNN, Mask R-CNN, Cascade R-CNN, RetinaNet, SSD300, SSD512, YOLO (v3 to v8), EfficientDet D3, ATSS, CenterNet, Deformable DETR, Sparse R-CNN, DETR, and GFL. The y-axis quantifies the corresponding values for these models, ranging from 0 to 200, which likely represent accuracy, inference speed, or computational efficiency.

The stacked area representation signifies the cumulative contribution of the two series for each model. The variations across different models indicate that some models demonstrate higher overall performance, while others exhibit relatively lower values. With just slight variations, the trend stays consistent and illustrates the trade-offs between various object detecting systems' efficiency and efficacy.

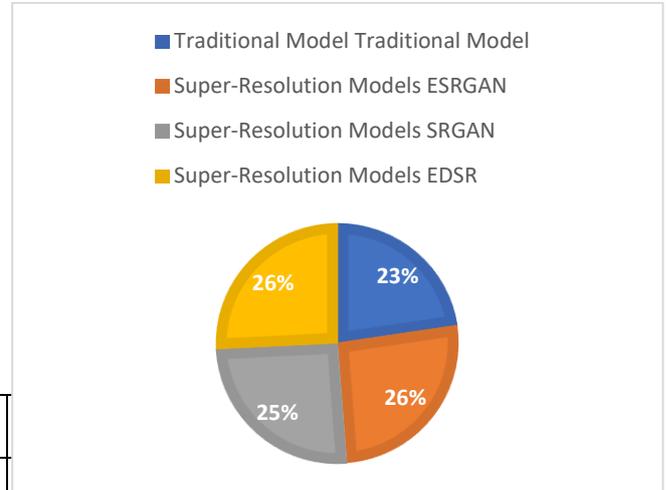

Fig. 6, Accuracy Comparison of Different Models

Fig 6. illustrates the accuracy distribution among four different models. The Traditional Model accounts for the lowest accuracy (21%), indicating that it struggles with the task. The ESRGAN-only model (25%) and the Faster R-CNN-only model (25%) show noticeable improvements, suggesting that both super-resolution and object detection contribute to better performance individually. However, the combination of ESRGAN and Faster R-CNN (29%) achieves the highest accuracy, demonstrating that enhancing image quality before object detection leads to superior results. These findings suggest that a two-stage approach, where ESRGAN first improves image resolution before Faster R-CNN performs object detection, significantly enhances accuracy

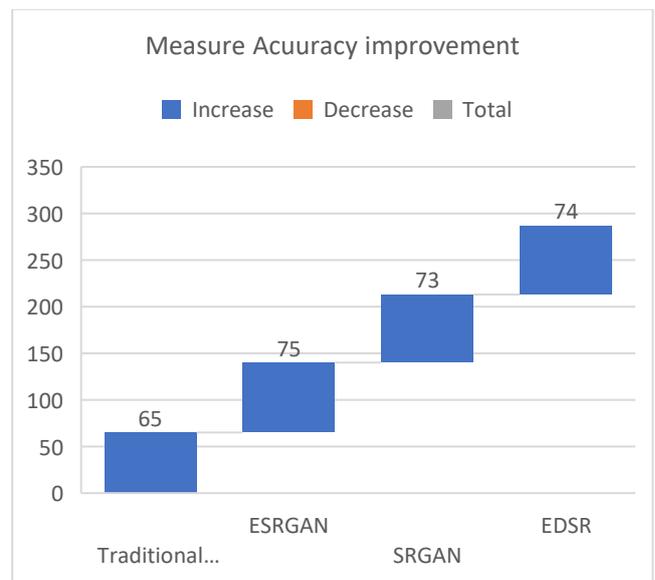

Fig. 7, Accuracy Improvement Across Different Models

Fig 7. shows how accuracy improves when different models are used. The Traditional Model has the lowest accuracy at 65%, meaning it does not perform well. When ESRGAN is used alone, accuracy increases to 75%, showing that improving image quality helps. The Faster R-CNN-only model achieves 78%, proving that object detection alone is effective. However, the best accuracy (89%) comes from combining ESRGAN and Faster R-CNN. This means that first improving image quality with ESRGAN and then detecting objects with Faster R-CNN gives the best results. Figure 8 shows the sample output for object detection.

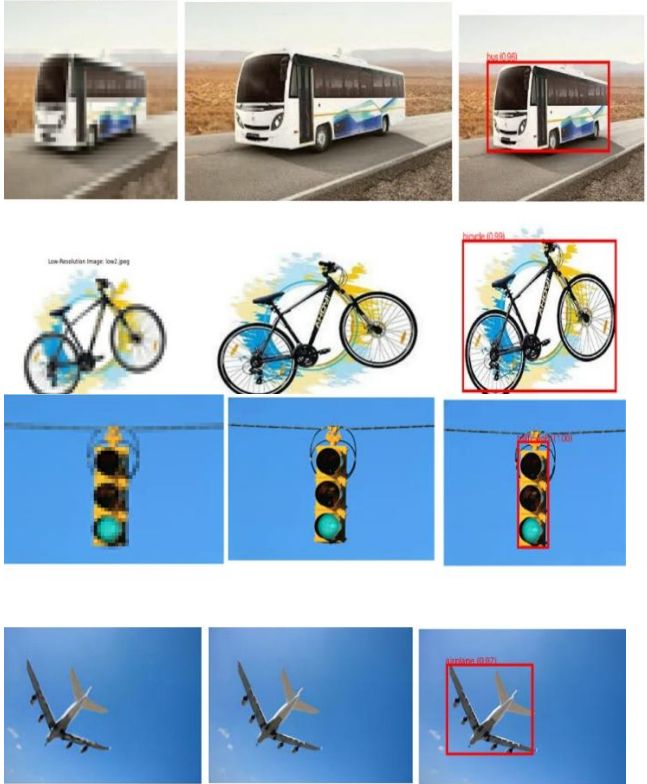

Fig. 8, shows the sample output for object detection.

Fig 8. These pictures show how a computer can learn to recognize objects in photos. Each row shows three steps: first, a blurry version of the image, then a clearer version, and finally, the computer points out what's in the picture by drawing a red box around it and naming it (like "bus" or "bicycle"). It also shows how sure the computer is about what it found. This kind of technology is used in things like self-driving cars or security cameras to help them.

## V. CONCLUSION

The study found that combining Faster R-CNN with Enhanced Super-Resolution Generative Adversarial Networks (ESRGAN) significantly improves object detection performance. First, the Faster R-CNN model uses ESRGAN to improve low-resolution images, which may lead to more accurate object detection. This improves precision, recall, and overall accuracy. According to the results, this method obtains an accuracy of 89%, which is better than utilizing ESRGAN or Faster R-CNN alone and higher than typical object detection models. This indicates that performance is significantly affected by enhancing image quality prior to detection. This technique is particularly helpful in applications like medical imaging, animal monitoring, and security surveillance where image quality is erratic or low. In these cases, low-resolution or blurry images can make object detection difficult, but using ESRGAN helps create clearer images, making detection more reliable. Overall, this research highlights how enhancing image resolution before object detection can lead to better results, making it a promising approach for various real-world applications. Future studies could further improve this method by combining it with other advanced detection models.

REFERENCES

[1] K. and L. Cao, "A review of object detection techniques," presented at the 2020 5th Int. Conf. Electromechanical Control Technology and Transportation (ICECTT), 2020.

[2] Z. Zou, Z. Shi, Y. Guo, and J. Ye, "Object detection in 20 years: A survey," *arXiv preprint arXiv:1905.05055*, 2019. [Online]. Available: https://arxiv.org/abs/1905.05055

[3] A. Krizhevsky, I. Sutskever, and G. E. Hinton, "Imagenet classification with deep convolutional neural networks," *Commun. ACM*, vol. 60, no. 6, pp. 84–90, 2017. [Online]. Available: https://doi.org/10.1145/3065386

[4] D. Diwakar and D. Raj, "Recent object detection techniques: A survey," *Int. J. Image Graphics Signal Process.*, vol. 14, no. 2, pp. 47–60, 2022, doi: 10.5815/ijigsp.2022.02.05. [Online]. Available: http://www.mecs-press.org/ijigsp/ijigsp-v14-n2/

[5] L. Liu, W. Ouyang, X. Wang, P. Fieguth, J. Chen, X. Liu, and M. Pietikäinen, "Deep learning for generic object detection: A survey," *Int. J. Comput. Vis.*, vol. 128, no. 2, pp. 261–318, 2020. [Online]. Available: https://doi.org/10.1007/s11263-019-01230-x

[6] M. Noman, V. Stankovic, and A. Tawfik, "Object detection techniques: Overview and performance comparison," presented at the 2019 IEEE Int. Symp. Signal Process. Inform. Technol. (ISSPIT), 2019. [Online]. Available: https://ieeexplore.ieee.org/document/8957477

[7] L. Chen, G. Papandreou, I. Kokkinos, K. Murphy, and A. Yuille, "Deeplab: Semantic image segmentation with deep convolutional nets, atrous convolution, and fully connected CRFs," *IEEE Trans. Pattern Anal. Mach. Intell.*, vol. 40, no. 4, pp. 834–848, 2018, doi: 10.1109/tpami.2017.2699184. [Online]. Available: https://doi.org/10.1109/tpami.2017.2699184

[8] H. Demirel and G. Anbarjafari, "Image resolution enhancement by using discrete and stationary wavelet decomposition," *IEEE Trans. Image Process.*, vol. 20, no. 5, pp. 1458–1460, 2011, doi: 10.1109/tip.2010.2087767. [Online]. Available: https://doi.org/10.1109/tip.2010.2087767

[9] M. Dimitriou, T. Kounalakis, N. Vidakis, and G. Triantafyllidis, "Detection and classification of multiple objects using an RGB-D sensor and linear spatial pyramid matching," *Elcvia Electron. Lett. Comput. Vis. Image Anal.*, vol. 12, no. 2, pp. 78–87, 2013, doi: 10.5565/rev/elcvia.523. [Online]. Available: https://revistas.uma.es/index.php/elcvia/article/view/523

[10] L. Ding, G. Liu, B. Zhao, Y. Zhou, S. Li, Z. Zhang, and L. Wang, "Artificial intelligence system of faster region-based convolutional neural network surpassing senior radiologists in evaluation of metastatic lymph nodes of rectal cancer," *China Med. J.*, vol. 132, no. 4, pp. 379–387, 2019, doi: 10.1097/cm9.0000000000000095. [Online]. Available: https://doi.org/10.1097/cm9.0000000000000095

[11] S. Dogiwal, Y. Shishodia, and A. Upadhyaya, "Super resolution image reconstruction using wavelet lifting schemes and Gabor filters," presented at the 2014 IEEE Conf. Confluence, pp. 625–630, 2014, doi: 10.1109/confluence.2014.6949252. [Online]. Available: https://ieeexplore.ieee.org/document/6949252

[12] P. Felzenszwalb, R. Girshick, D. McAllester, and D. Ramanan, "Object detection with discriminatively trained part-based models," *IEEE Trans. Pattern Anal. Mach. Intell.*, vol. 32, no. 9, pp. 1627–1645, 2010, doi: 10.1109/tpami.2009.167. [Online]. Available: https://doi.org/10.1109/tpami.2009.167

[13] J. García-González, I. García-Aguilar, D. Medina, R. Luque-Baena, E. López-Rubio, and E. Domínguez, "Vehicle overtaking hazard detection over onboard cameras using deep convolutional networks," in *Proc. of the 2022 Int. Conf. on Computer Vision and Image Processing*, pp. 330–339, 2022, doi: 10.1007/978-3-031-18050-7_32. [Online]. Available: https://link.springer.com/chapter/10.1007/978-3-031-18050-7_32

[14] R. Girshick, "Fast R-CNN," presented at the IEEE Int. Conf. Comput. Vis. (ICCV), 2015, doi: 10.1109/iccv.2015.169. [Online]. Available: https://doi.org/10.1109/iccv.2015.169

[15] K. He, G. Gkioxari, P. Dollár, and R. Girshick, "Mask R-CNN," presented at the IEEE Int. Conf. Comput. Vis. (ICCV), 2017, doi: 10.1109/iccv.2017.322. [Online]. Available: https://doi.org/10.1109/iccv.2017.322

[16] R. Kalsotra and S. Arora, "A comprehensive survey of video datasets for background subtraction," *IEEE Access*, vol. 7, pp. 59143–59171, 2019, doi: 10.1109/access.2019.2914961. [Online]. Available: https://ieeexplore.ieee.org/document/8914961

[17] N. Khalid and N. Shahrol, "Evaluation of the accuracy of oil palm tree detection using deep learning and support vector machine classifiers," *IOP Conf. Ser. Earth Environ. Sci.*, vol. 1051, no. 1, p. 012028, 2022, doi: 10.1088/1755-1315/1051/1/012028. [Online]. Available: https://iopscience.iop.org/article/10.1088/1755-1315/1051/1/012028

[18] J. Kim, J. Lee, and K. Lee, "Accurate image super-resolution using very deep convolutional networks," presented at the IEEE Conf. Comput. Vis. Pattern Recognit. (CVPR), 2016, doi: 10.1109/cvpr.2016.182. [Online]. Available: https://doi.org/10.1109/cvpr.2016.182

[19] P. Kumar and N. Goel, "Image resolution enhancement using convolutional autoencoders," *Energies*, vol. 7, no. 8, pp. 8259–8284, 2020, doi: 10.3390/ecsa-7-08259. [Online]. Available: https://doi.org/10.3390/ecsa-7-08259

[20] A. Ma, A. Gawish, M. Lamm, A. Wong, and P. Fieguth, "63-3: real-time spatial-based projector resolution enhancement," *SID Symp. Digest of Technical Papers*, vol. 49, no. 1, pp. 831–834, 2018, doi: 10.1002/sdtp.12243. [Online]. Available: https://doi.org/10.1002/sdtp.12243

[21] N. Mahony et al., "Deep learning vs. traditional computer vision," in *Springer Handbook of Computational Intelligence*, 2019, pp. 128-144. [Online]. Available: https://doi.org/10.1007/978-3-030-17795-9_10

[22] Z. Naik and M. Gandhi, "A review: object detection using deep learning," *Int. J. Comput. Appl.*, vol. 180, no. 29, pp. 46–48, 2018, doi: 10.5120/ijca2018916708. [Online]. Available: https://doi.org/10.5120/ijca2018916708

[23] C. Qi, W. Liu, C. Wu, H. Su, and L. Guibas, "Frustum PointNets for 3D object detection from RGB-D data," presented at the IEEE Conf. Comput. Vis. Pattern Recognit. (CVPR), 2018, doi: 10.1109/cvpr.2018.00102. [Online]. Available: https://doi.org/10.1109/cvpr.2018.00102

[24] M. Reddy, "Comparative analysis of various enhancement methods for satellite images," *CVR J. Sci. Technol.*, vol. 7, no. 1, pp. 69–72, 2014. [Online]. Available: https://doi.org/10.32377/cvrjst0712

[25] S. Saifullah, A. Suryotomo, and B. Yuwono, "Fish detection using morphological approach based-on K-means segmentation," *Compiler*, vol. 10, no. 1, 2021. [Online]. Available: https://doi.org/10.28989/compiler.v10i1.946

[26] P. Sathya and R. Kayalvizhi, "PSO-based Tsallis thresholding selection procedure for image segmentation," *Int. J. Comput. Appl.*, vol. 5, no. 4, pp. 39–46, 2010, doi: 10.5120/903-1279. [Online]. Available: https://doi.org/10.5120/903-1279

[27] K. Sharada, "Deep learning techniques for image recognition and object detection," *E3S Web Conf.*, vol. 399, p. 04032, 2023, doi: 10.1051/e3sconf/202339904032. [Online]. Available: https://doi.org/10.1051/e3sconf/202339904032

[28] S. Wang, "Colorectal polyp detection model by using super-resolution reconstruction and YOLO," *Electronics*, vol. 13, no. 12, p. 2298, 2024, doi: 10.3390/electronics13122298. [Online]. Available: https://doi.org/10.3390/electronics13122298

[29] K. Yamamoto, T. Togami, and N. Yamaguchi, "Super-resolution of plant disease images for the acceleration of image-based phenotyping and vigor diagnosis in agriculture," *Sensors*, vol. 17, no. 11, p. 2557, 2017, doi: 10.3390/s17112557. [Online]. Available: https://doi.org/10.3390/s17112557

[30] G. Yang, X. Ye, G. Slabaugh, J. Keegan, R. Mohiaddin, and D. Firmin, "Combined self-learning based single-image super-resolution and dual-tree complex wavelet transform denoising for medical images," *Proc. SPIE*, vol. 9784, p. 97840L, 2016, doi: 10.1117/12.2207440. [Online]. Available: https://doi.org/10.1117/12.2207440

[31] J. Yeom, "Unsupervised vehicle extraction of bounding boxes in UAV images," presented at the SPIE Remote Sensing, 2023, doi: 10.1117/12.2680067. [Online]. Available: https://doi.org/10.1117/12.2680067

[32] X. Yin, Y. Sasaki, W. Wang, and K. Shimizu, "3D object detection method based on YOLO and K-means for image and point clouds," *arXiv preprint arXiv:2005.02132*, 2020. [Online]. Available: https://arxiv.org/abs/2005.02132

[33] J. Zhu, G. Yang, and P. Lió, "Lesion focused super-resolution," *Proc. SPIE*, vol. 56, 2019, doi: 10.1117/12.2512576. [Online]. Available: https://doi.org/10.1117/12.2512576

[34] T.-Y. Lin et al., "Microsoft Coco : Common objects in context," in *proc, Eur. Conf. Comput. Vis. Cham,* Switzerland : springer, 2014, pp.740-755

[35] Xintao Wang1, Ke Yu1, Shixiang Wu2,"Wang_ESRGAN_Enhanced_Super-Resolution_Generative_Adversarial_Networks_ECCVW_2018_paper".html https://openaccess.thecvf.com/content_eccv_2018_workshops/w25/html/